%% file: manuscript.tex
\documentclass{article}
\usepackage[utf8]{inputenc}
\usepackage[margin=0.9in]{geometry}

\usepackage{booktabs}
\usepackage{amsfonts}
\usepackage{amsmath}
\usepackage{amssymb}
\usepackage{amsthm}
\usepackage{mathtools}
\usepackage{nicematrix}

\usepackage{rotating}
\usepackage{graphicx}
\usepackage{subfigure}
\usepackage{color}
\usepackage{tikz-cd}
\usetikzlibrary{arrows,backgrounds}
\usepackage{verbatim}
\usepackage{hyperref}
\usepackage{bm}

\usepackage[ruled,vlined]{algorithm2e}

\usepackage{cleveref}

\crefname{algocf}{algorithm}{algorithms}
\Crefname{algocf}{Algorithm}{Algorithms}

\usepackage{thmtools}
\usepackage{thm-restate}

\newcommand{\norm}[1]{\left \lVert #1 \right \rVert}

\title{Domain-Decomposed Graph Neural Network \\Surrogate Modeling for Ice Sheets}
\author{Adrienne M. Propp\thanks{Institute for Computational and Mathematical Engineering, Stanford University, propp@stanford.edu} \and Mauro Perego \thanks{Department of Scientific Machine Learning, Sandia National Laboratories, mperego@sandia.gov} \and Eric C. Cyr \thanks{Department of Scientific Machine Learning, Sandia National Laboratories, eccyr@sandia.gov} \and Anthony Gruber \thanks{Department of Scientific Machine Learning, Sandia National Laboratories, adgrube@sandia.gov} \and Amanda Howard \thanks{Advanced Computing, Mathematics and Data Division, Pacific Northwest National Laboratory} \and Alexander Heinlein \thanks{Delft Institute of Applied Mathematics, Delft University of Technology, Delft, Netherlands, a.heinlein@tudelft.nl} \and Panos Stinis \thanks{Advanced Computing, Mathematics and Data Division, Pacific Northwest National Laboratory} \and Daniel Tartakovsky \thanks{Department of Energy Science Engineering, Stanford University}}

\begin{document}
\maketitle

\begin{abstract}
Accurate yet efficient surrogate models are essential for large-scale simulations of partial differential equations (PDEs), particularly for uncertainty quantification  (UQ) tasks that demand hundreds or thousands of evaluations. We develop a physics-inspired graph neural network (GNN) surrogate that operates directly on unstructured meshes and leverages the flexibility of graph attention. To improve both training efficiency and generalization properties of the model, we introduce a domain decomposition (DD) strategy that partitions the mesh into subdomains, trains local GNN surrogates in parallel, and aggregates
their predictions. We then employ transfer learning to fine-tune models across subdomains, accelerating training and improving accuracy in data-limited settings. Applied to ice sheet simulations, our approach accurately predicts full-field velocities on high-resolution meshes, substantially reduces training time relative to a single global surrogate model, and provides a ripe foundation for UQ objectives. Our results demonstrate that graph-based DD, combined with transfer learning, provides a scalable and reliable pathway for training GNN surrogates on massive PDE-governed systems, with broad potential for application beyond ice sheet dynamics.
\end{abstract}

\input{Sections/S1_intro}
\input{Sections/S2_ice_model}
\input{Sections/S3_computational_approach}
\input{Sections/S4_results}
\input{Sections/S5_conclusion}

\clearpage
\newpage
\appendix
\input{Sections/S6_appendix}

\bibliographystyle{siamplain}
\bibliography{references}

\end{document}

%% file: Sections/S1_intro.tex
\section{Introduction}
Many important scientific phenomena are governed by complex, nonlinear partial differential equations (PDEs) posed on intricate, evolving geometries. High-fidelity numerical solvers for such systems, while accurate, are often prohibitively slow: a single simulation may take hours, days, or longer. Decision-oriented tasks such as uncertainty quantification (UQ), which require hundreds or thousands of model evaluations, can therefore become computationally intractable using traditional solvers alone. This has created a growing demand for efficient modeling techniques that not only accelerate simulations but also respect the underlying physics, scale to massive problems, and generalize across changes in mesh resolution, input parameters, and domain geometry.

In the present work, we address these challenges by developing a physics-inspired graph neural network (GNN)-based surrogate model, applied to the case of ice sheets. Ice sheet dynamics provide a compelling and high-impact testbed. These simulations involve coupled PDEs on intricate domains, representing vast regions across which physical processes vary significantly.  Reliable predictions of ice movement and mass loss underpin critical policy decisions concerning coastal infrastructure, resource planning, and climate adaptation. However, the computational cost of traditional methods often precludes comprehensive UQ, limiting our ability to quantify risk and explore plausible scenarios. Efficient and accurate surrogate models are therefore essential in this domain.

While recent advances in machine learning (ML) offer several promising directions for data-driven surrogate modeling, GNNs are naturally suited to the setting of modeling physical systems on unstructured meshes. MeshGraphNets \cite{pfaff2020learning} was among the first to demonstrate that GNNs can accurately learn physical dynamics on mesh-based PDE simulations while operating directly on irregular meshes and without requiring remeshing or post-processing. Alternative approaches, such as operator learning via DeepONets \cite{he2023hybrid}, have also shown strong performance for ice sheet modeling. However, because both the branch and trunk networks are tied to fixed sensor locations and domain geometry, standard DeepONets but must be retrained for each change in the computational domain and mesh, limiting their generalization and flexibility.

However, training surrogates at the scale of ice sheets, where tens of thousands of nodes represent millions of square kilometers, requires innovative techniques to ensure accuracy, generalization and computational efficiency. Transfer learning has emerged as a powerful tool in data-limited regimes \cite{howard2024multifidelity,li2024adaptergnn,propp2025transfer}, allowing models to be pre-trained on broad distributions and then fine-tuned on smaller, task-specific datasets. Separately, domain decomposition (DD) methods have long been a cornerstone of classical numerical simulations, reducing computational and memory costs as well as enabling concurrency of computation by partitioning the domain into smaller subdomains before solving \cite{dolean_introduction_2015,quarteroni_domain_1999,smith_domain_2004,toselli_domain_2005}. These ideas are now being revisited in the context of ML surrogates \cite{heinlein_combining_2021,klawonn_machine_2024}, though they have not yet been extended to the case of GNNs. In our work, we combine transfer learning and domain decomposition to obtain major improvements in convergence speed, accuracy, and computational efficiency in our GNN surrogate.

While the modeling framework we propose has far-reaching implications beyond the context of ice sheet dynamics, we focus on predicting ice sheet velocity at a given time $t$ from basal friction, bed topography (or elevation), and thickness at time $t$. Our main contributions include:
\begin{enumerate}
    \item A physics-inspired GNN surrogate that accurately models ice sheet velocity on a large, unstructured mesh, 
    \item A transfer learning strategy that accelerates GNN training, and     \item A domain decomposition (DD) framework for GNNs.
\end{enumerate}

Together, these contributions advance the development of efficient, accurate graph-based surrogate models for large-scale physical systems, and lay the groundwork for future UQ studies of ice sheet dynamics and other systems.

The remainder of the paper is organized as follows. \Cref{sec:ice_sheet_model} introduces the ice sheet model and governing equations. \Cref{sec:computational_approach} describes our computational approach, including an overview of GNNs, our specific architecture, and our training setup. \Cref{sec:TL,sec:DD} introduce our transfer learning and domain decomposition strategies, respectively. \Cref{sec:results} presents results from our computational experiments, and \Cref{sec:discussion} concludes with a discussion of implications and future directions, particularly for uncertainty quantification.

%% file: Sections/S2_ice_model.tex
\section{Physics-based model and data generation} \label{sec:ice_sheet_model}

As a case study, we demonstrate our surrogate modeling approach on Greenland's Humboldt Glacier. Spanning nearly 2 million square kilometers, Humboldt Glacier is one of Greenland's largest glaciers. Its massive size and complex dynamics make it an ideal test case for large-scale surrogate modeling of physical systems. 

\begin{figure}[hbt!]
\centering
\includegraphics[width=0.7\linewidth]{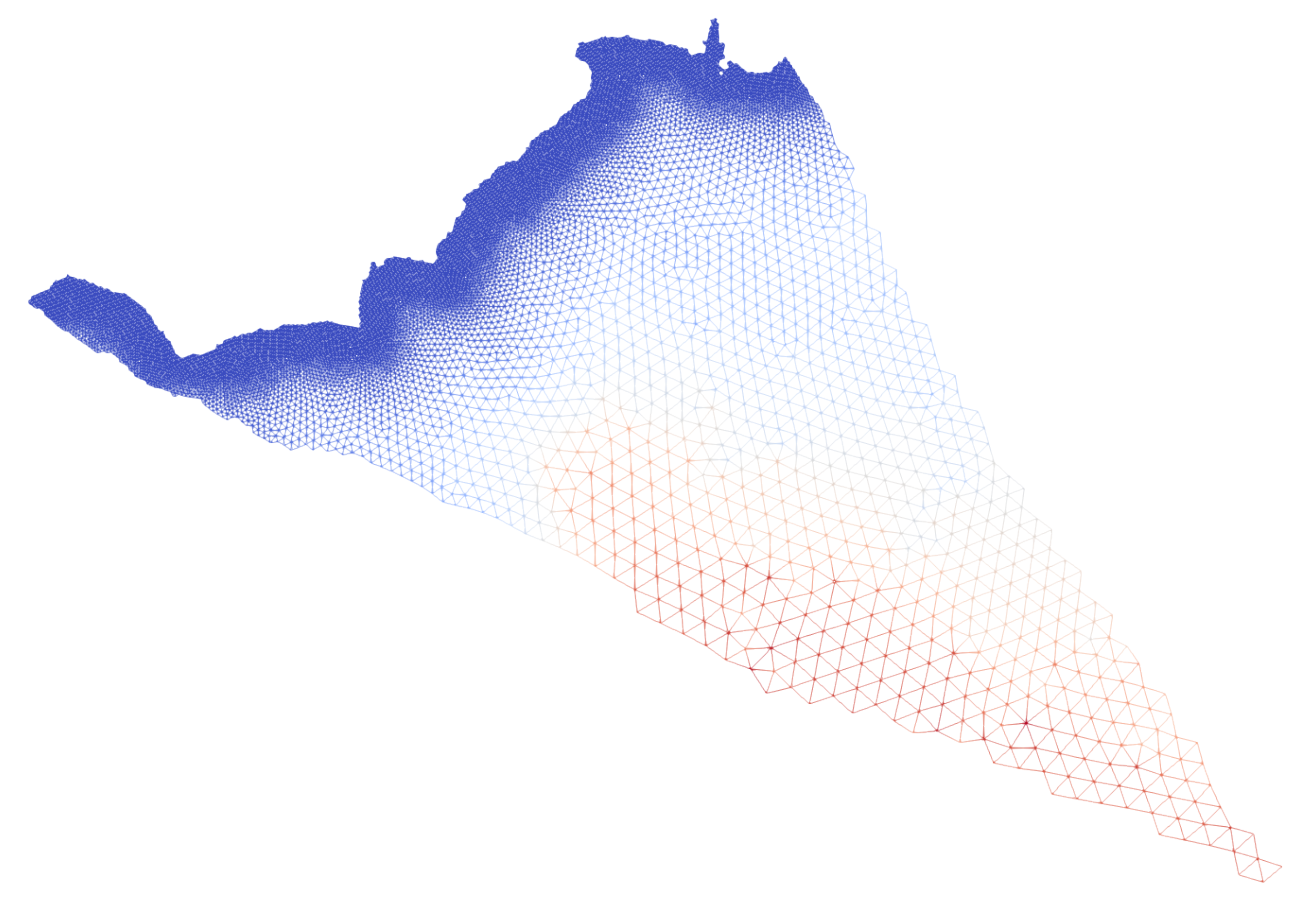}
\caption{Dual Delaunay triangulation used as the graph representation of the Humboldt Glacier. In the MALI ice sheet model, ice thickness and temperature are discretized using a finite-volume method on a Voronoi mesh, while the velocity equations are solved on its dual Delaunay triangulation using low-order Lagrangian finite elements. We use this dual mesh as the node–edge structure for our GNN surrogate. Colors indicate cell area, with red indicating large cell area and blue indicating small cell area. Note that cell area decreases toward the glacier terminus where velocities and uncertainty are greatest, corresponding to finer mesh resolution in the finite element simulations.
}
\label{fig:graph}
\end{figure}

We use the MPAS-Albany Land Ice (MALI) ice-sheet model \cite{hoffman2018mpas} to simulate glacier dynamics, governed by the equations of ice flow described below. MALI solves the governing equations using a finite-volume and finite-element discretization on unstructured meshes that are adaptively refined where higher accuracy is needed (see \Cref{fig:graph}). In our setup, the ice thickness and temperature equations are solved using a finite volume method with explicit forward-Euler time stepping on a Voronoi mesh of the Humboldt Glacier, while the velocity equations are solved on its dual Delaunay triangulation using low-order Lagrangian finite elements. Below, we briefly summarize the key equations and constitutive relationships governing ice thickness and velocity evolution. We refer the interested reader to \cite{hillebrand2022} for additional details on the governing equations (and topics such as the temperature model and calving), and to \cite{hoffman2018mpas,watkins2023performance} for additional details on the Humboldt dataset setup.

\subsection{Ice sheet model equations} \label{sec:MALI}
The evolution of ice thickness $H(x,y,t)$ is governed by the continuity equation:
\begin{align}
    \partial_t H+\nabla\cdot(\overline{\mathbf{u}}H)=a_H,
\end{align}
where
\begin{align}
\overline{\mathbf{u}}=\frac{1}{H}\int_l^s\mathbf{u}dz
\end{align}
is the depth-integrated velocity integrated between the glacier bed $z=l(x,y,t)$ and upper surface $z=s(x,y,t)$. The accumulation-ablation term $a_H$ accounts for surface and basal mass balance, including both snowfall and melting. We adopt the convention that $z=0$ corresponds to sea level, and define the bedrock surface $b(x,y)$ such that $l(x,y,t)=b(x,y)$ for grounded ice and $l(x,y,t)=-\frac{\rho}{\rho_w}H(x,y,t)$ for floating ice, where $\rho$ and $\rho_w$ are the densities of ice and seawater, respectively.

At the ice-sheet scale, ice behaves as an incompressible, shear-thinning fluid, with velocity $\mathbf{u}(x,y,z)$ satisfying Blatter-Pattyn approximation~\cite{blatter1995velocity,pattyn2003new}. This model simplifies the full nonlinear Stokes equations for shallow ice sheets by neglecting small terms in the vertical momentum balance and strain-rate tensor, yielding
\begin{equation}
    -\nabla \cdot (2\eta(\mathbf u)\, \mathbf{D(\mathbf u)} ) = - \rho g \nabla s.
\end{equation}
Here, $\rho$ is the ice density, $g$ is acceleration due to gravity, and $\eta$ is the effective viscosity of ice, defined as:
\begin{equation} \label{viscosity}
    \eta = \frac12 A(T)^{-q} \, D_e(\mathbf u)^{q-1},
\end{equation}
with effective strain rate
\begin{equation}
    D_e(\mathbf u) = \sqrt{u_x^2 + v_y^2 + u_x v_y + \frac{1}{4} \big(u_y + v_x\big)^2 + \frac{1}{4} \big(u_z^2 +v_z^2}\big).
\end{equation}
In \eqref{viscosity}, $A(T)$ is the temperature-dependent rate factor. We assume $q=1/3$, which is standard practice and corresponds to Glen's flow law exponent $n=3$ \cite{cuffey2010physics,glen1955creep}. The modified strain-rate tensor $\mathbf{D}(\mathbf u)$ is defined as:
\begin{equation}
    \mathbf{D}(\mathbf u) = \begin{bmatrix} 2u_x + v_y & \frac12 (u_y + v_x) & \frac12 u_z \\
    \frac12 (u_y + v_x) & u_x + 2v_y & \frac12 v_z \end{bmatrix},
\end{equation}
where $u$ and $v$ are the horizontal components of the velocity $\mathbf u$, and subscripts denote partial derivatives. Note that the Blatter-Pattyn model does not solve for the vertical component of the velocity, but it can be recovered from the the incompressibility condition $\nabla\cdot\mathbf{u}=0$.
Also, note that viscosity depends on the temperature $T$, which satisfies a heat equation (see \cite{hoffman2018mpas} for more information).

Boundary conditions, particularly those governing the sliding of ice along the glacier bed, are of primary importance in ice sheet modeling. In this work, we model basal sliding using Budd's law:
\begin{equation}
2 \eta(\mathbf u) \mathbf{D}(\mathbf u) \mathbf{n} = \mu N |\mathbf u|^{q-1} \mathbf{u},
\end{equation}
where $N= \max(\rho g H - \rho_w g z, 0)$ is the effective pressure at the bed, and $\mu(x,y)$ is a spatially varying basal friction coefficient. The field $\mu$ is a major source of uncertainty in ice sheet models, as it cannot be measured directly. Quantifying uncertainty in $\mu$ is therefore essential for producing reliable projections of glacial dynamics. Accelerating this analysis is one of the central motivations of this work. In the next section, we discuss how to approximate the probability distribution of $\mu$.

We adopt the Mono-Layer Higher-Order (MOLHO) approximation \cite{dias2022new}, as implemented in MALI \cite{hillebrand2025antarctica}, which solves the Galerkin weak form of the Blatter-Pattyn (or higher-order) equations using a depth-separable velocity ansatz:
\begin{equation}
\mathbf u(x,y,z) = \Phi(z) \mathbf u_b(x,y) + \left(1-\Phi(z)\right) \mathbf u_s(x,y),
\end{equation}
where
\begin{equation}
    \Phi(z) = \left(\dfrac{s-z}{H}\right)^{\frac{1}{q}+1}.
\end{equation}
Here, $\mathbf{u_b}$ and $\mathbf{u_s}$ denote velocities at the bed and upper surface of the ice-sheet, respectively. The shape functions $\Phi(z)$ and $\left(1-\Phi(z)\right)$ are also used to define the test functions of the weak formulation.

\subsection{Basal Friction Parameter Distribution} \label{sec:posterior}
The basal friction field $\mu(x,y)$ plays a central role in ice sheet dynamics but cannot be observed directly. In practice, $\mu$ is typically estimated by solving a PDE-constrained optimization problem \cite{Goldberg_2015,MacAyeal_1993, Morlighem_2010, Perego_PS_JGRES_2014,Petra_2012} to identify the basal friction field that minimizes the misfit between observed surface velocities and those predicted by the physical model. To ensure positivity, we work with the transformed variable $p = \log (\mu)$. 

We seek a probability distribution for $p$ 
that reflects uncertainties in  the observations, model, and estimated parameter itself. Classical Bayesian inference methods, such as Markov Chain Monte Carlo (MCMC), become intractable for such high-dimensional inverse problems. Instead, we adopt the Laplace approximation, which has been successfully applied in ice sheet modeling \cite{brinkerhoff_2025_forecast,jakeman_2025_multifidelity,Recinos_GMT_TC_2023} after being first proposed in this context by \cite{Isaac_PSG_JCP_2015}. The Laplace approximation constructs a quadratic (Gaussian) approximation of the log-posterior distribution in a neighborhood of the maximum a posteriori (MAP) estimate $p_\text{MAP}$, which can be obtained from the PDE-constrained optimization approach mentioned earlier. Although computing the MAP point is itself nontrivial, the resulting Gaussian approximation provides a tractable and scalable representation of posterior uncertainty.

Under this approximation, the posterior distribution of $p$ is
\begin{equation}
    p\sim\mathcal{N} (p_\text{MAP}, \Sigma^\text{post}).
\end{equation} Samples can then be generated as:
\begin{equation}
p = p_{\text{MAP}} + L \omega,  \quad \omega \sim \mathcal N(0,I),
\end{equation}
where $L$ satisfies $L L^T = \Sigma^\text{post}$. In practice, neither $L$ nor $\Sigma^\text{post}$ is computed explicitly. Instead, we compute the matrix-free operation $L \omega$, allowing sampling from the posterior without assembling dense covariance matrices. The corresponding basal friction realizations are then obtained as $\mu = \exp(p)$. Further details on this approach can be found in \cite{Isaac_PSG_JCP_2015}.

We adopt a Gaussian prior on $p$ of the form: \begin{equation}
    p\sim\mathcal N(0, \mathcal A^{-2}),
\end{equation}
where $\mathcal A$ is a precision operator, the infinite-dimensional analogue of a precision matrix (or inverse covariance matrix) in finite-dimensional Gaussian distributions. In PDE-based priors, the precision operator is a differential operator that imposes smoothness. Specifically, we define $\mathcal{A}$ as:
\begin{equation}
\mathcal A := \left \{ \begin{array}{ll}
-\gamma \Delta p + \delta p & \text{in } \Gamma_b, \\
-\gamma \nabla p \cdot \mathbf{n} -\xi p & \text{on } \partial \Gamma_b. \\
\end{array} \right .
\end{equation}
This represents a second-order elliptic operator with Neumann-type boundary conditions, promoting smooth friction fields while permitting sharp variations where supported by data. The notation $\mathcal A^{-2}$ denotes the corresponding covariance operator, a convention common in PDE-constrained inverse problems. This construction yields a prior with the desired Sobolev regularity and spatial correlation structure.
We set $\gamma = 8.976$~km$^2$, $\delta = 8.865\times 10^{-3}$, and $\xi=0.1987$~km$^{-1}$, yielding a prior with approximately unit marginal variance and a correlation length of $90$km (see \cite{villa2024prior}).

Observational data enter through the likelihood term, which depends on the observed surface velocity $\mathbf{u}_s$. We assume $\mathbf u_s^{\text{obs}}$ is observed with covariance $\Sigma^{\text{obs}}$, inflated to account for both measurement uncertainty and model error. This setup allows us to generate ensembles of basal friction fields that capture physically plausible variability and propagate this uncertainty into the resulting ice flow simulations.

\subsection{Data generation}
To generate training data for our surrogate model, we sample basal friction fields $\mu(x,y)$ from the posterior distribution described in \Cref{sec:posterior} and evolve the ice sheet forward from year 2007 to year 2100 using the MALI ice sheet model.
Our setup builds on earlier work, e.g. \cite{he2023hybrid}, but incorporates several key improvements that provide more realistic physics and substantially richer training data.

First, the MALI ice sheet model used here includes physical processes absent from \cite{he2023hybrid}, including calving, basal melting, and temperature evolution. Second, we adopt Budd's nonlinear sliding law, which is more realistic than the linearized form used previously. Third, the Humboldt Glacier mesh resolution is roughly eight times finer than in \cite{he2023hybrid}, allowing for the representation of steep gradients near the terminus.  Finally, our basal friction probability distribution is constrained by both data and physics, while \cite{he2023hybrid} imposed a simple squared-exponential Gaussian process prior. Together, these advances yield substantially higher-fidelity simulation data and provide a more challenging and representative test case for surrogate modeling.

\begin{figure}[hbt!]
\centering
\includegraphics[width=\linewidth]{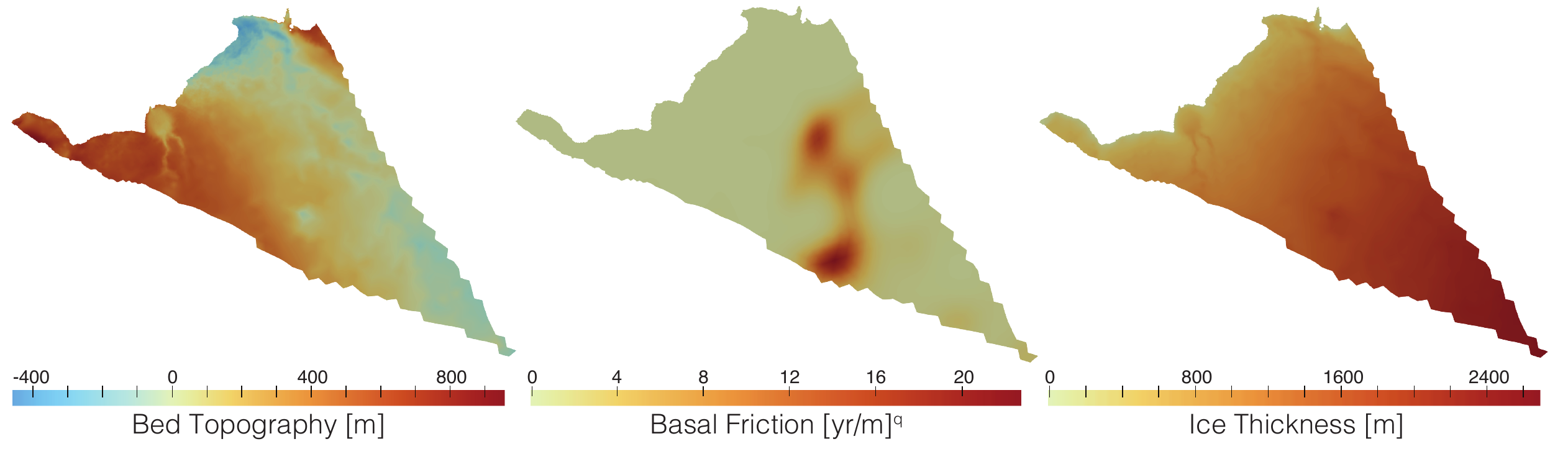}
\caption{Sample realizations of the input features for our GNN surrogate, plotted on the Humboldt Glacier. Left: bed topography [m]; Center: basal friction [yr/m]$^{q}$, where the value of $q$ accounts for the nonlinearity of the sliding law; Right: ice thickness [m].}
\label{fig:input_features}
\end{figure}

%% file: Sections/S3_computational_approach.tex
\section{Computational approach}\label{sec:computational_approach}
In this section, we describe the components of our surrogate modeling framework, including an overview of graph neural networks (GNNs), the specific bracket-based GNN architecture we adopt, the construction of the computational graph, our training protocol, and the fine-tuning and domain-decomposition strategies we used to improve training efficiency and generalization. Throughout, we emphasize the design choices motivated by the physics of ice-sheet flow and the limitations of standard message-passing GNNs when applied to large-scale scientific simulations.

\input{Sections/S3_p1_GNN}
\input{Sections/S3_p2_BracketGraphs}
\input{Sections/S3_p3_dataprep}

\input{Sections/S3_p4_training}
\input{Sections/S3_p5_transfer_learning}

\input{Sections/S3_p6_domain_decomposition}

%% file: Sections/S3_p1_GNN.tex
\subsection{Graph neural networks (GNNs)}\label{sec:GNNs}
Many problems in computational physics, including ice-sheet modeling, involve solving PDEs on irregular geometries and unstructured meshes. Most standard ML architectures assume that data lie on a regular, Euclidean grid. Applying them to unstructured FEM meshes requires remeshing or interpolation, which can introduce error, lose resolution, and increase computational cost. GNNs offer a natural alternative: they are designed for relational data, support permutation invariance, and operate directly on unstructured meshes \cite{wu2020comprehensive}. These properties make GNNs a promising tool for surrogate modeling of ice dynamics and other physical systems \cite{pfaff2020learning,shukla2022scalable}, where geometric flexibility and the ability to incorporate new meshes or subdomains are important.

The message-passing GNN modeling framework, proposed in \cite{gilmer2017neural}, represents each node with a feature vector containing physical variables such as ice thickness and velocity. Although message-passing GNNs can perform predictions at the graph, edge, or node level, those designed for physical simulations primarily conduct node-level prediction through iterative aggregation and feature transformation operations on neighboring nodes.
A typical GNN layer updates the node features according to:
\begin{align}
    h_i^{(l+1)}=\sigma\bigg(W^{(l)}\cdot\text{AGG}\big(\{h_j^{(l)}:j\in\mathcal{N}(i)\}\cup\{h_i^{(l)}\}\big)\bigg),
\end{align}
where $h_i^{(l)}$ denotes the representation of node $i$ at layer $l$, $\mathcal{N}(i)$ denotes the neighborhood of node $i$, $W^{(l)}$ is a matrix of learnable weights, $\text{AGG}$ is an aggregation operator (typically sum or mean), and $\sigma$ is a nonlinear activation function. This framework enables the network to capture localized spatial interactions while aggregating broader contextual information across multiple layers. A reframing of this algorithm in the context of sparse linear algebra is detailed in \cite{moore2025graph}.

For large-scale domains with heterogeneous physical behavior, GNNs offer several important computational advantages. GNNs perform learning through local message-passing operations that are applied uniformly across all nodes and edges. Because the same learned functions operate on node and edge features regardless of the underlying connectivity, model parameters can be transferred seamlessly across different graphs, as long as the feature dimensions are consistent. This enables flexible training strategies such as training on subgraphs, fine-tuning on related regions, and incorporating new data even when it arises from a different mesh or a modified domain geometry (see \Cref{sec:DD} for our DD approach). This flexibility is difficult to achieve with other deep learning architectures. For example, the DeepONet model developed in \cite{he2023hybrid} must be re-trained for each new ice sheet or mesh configuration, limiting its usefulness.

Despite these advantages and successes across diverse modeling tasks (e.g.,\cite{Koo_Rahnemoonfar_2025,lam_learning_2023,merchant_scaling_2023,stokes_deep_2020}), GNNs suffer from the notable drawback known as oversmoothing, where repeated averaging causes node features to become indistinguishable after only a small number of layers \cite{li_deeper_insights,rusch2023survey,wu2023a, wu2020comprehensive}. 
Oversmoothing arises from the spectral properties of the graph Laplacian and the diffusive nature of standard message passing operators \cite{cai_note_2020,li_deeper_insights}. As information propagates through successive layers, high-frequency components of the signal are dampened. While the smoothing of node features can be advantageous for applications like node classification\footnote{This is because classification tasks can be easier when the features of nodes in the same cluster are more similar.} \cite{li_deeper_insights}, it ultimately leads to a loss of fine-grained spatial detail essential for capturing complex physical phenomena. For problems like ice sheet modeling, where preserving sharp gradients in quantities such as ice thickness, basal friction, and velocity fields is essential for accurate predictions, this presents a major challenge.

Various mitigations for oversmoothing have been proposed: for example, adding skip connections (such as the ``multimesh'' edge hierarchy introduced by \cite{lam_learning_2023}) to propagate long-range information with fewer message-passing steps; group normalization and stochastic edge-dropping, permitting GNNs to go deeper in node classification tasks \cite{zhou_twards_deeper}; and architectures that incorporate physically informed inductive biases to explicitly separate conservative and diffusive processes \cite{bishnoi_enhancing_2024,sanchez-gonzalez_hamiltonian_2019}. We build on this last line of work by adopting the bracket-based GNN architecture of \cite{gruber2023reversible}, which reformulates message passing through the lens of bracket-based partial differential equations. This physics-inspired framework mitigates
oversmoothing and provides a principled mechanism for incorporating physical constraints and conservation laws
directly into the GNN architecture. This ultimately yields more physically consistent and stable surrogate models for
physical problems like ice sheet dynamics.

%% file: Sections/S3_p2_BracketGraphs.tex
\subsection{Bracket-based GNN architecture}\label{sec:BracketGraphs}

We use the Hamiltonian bracket-based GNN architecture introduced in \cite{gruber2023reversible} as the core of our surrogate model, adapting it to the setting of large-scale physics simulations. Key to this is a  reformulation of GNN message passing as a Hamiltonian dynamical system, where information propagates 
according to energy-conserving dynamics rather than diffusive averaging operations. This approach naturally avoids feature oversmoothing by preventing homogenization during message passing.

At a high level, the architecture processes information in three phases:
\begin{enumerate}
    \item \textbf{Encoding phase}. Raw node-edge feature pairs are lifted to a higher-dimensional latent space: $(\mathbf{q}',\mathbf{p}') = \mathbf{x}' \mapsto  \mathbf{x}=(\mathbf{q},\mathbf{p})$;
    \item \textbf{Physics-inspired message-passing phase}. Latent features $\mathbf{x}$ evolve to pseudo-time $T>0$ (analogous to depth)  according to an autonomous graph neural ODE:
    \begin{equation}
    \dot{\mathbf{x}}=F_{\theta}(\mathbf{x}),
    \end{equation}
    where $F_{\theta}$ is 
    constrained to generate a bracket-based dynamical system. Depending on the choice of bracket, this yields conservative, dissipative, or metriplectic flows, providing guarantees on stability and preventing oversmoothing.
    \item \textbf{Decoding phase}. The final latent state $\mathbf{x}(T)$ is mapped back to physical space to producing node-level predictions.
\end{enumerate} 

The key differentiator between this architecture and standard GNNs lies in its message-passing phase: instead of a discrete stack of message-passing layers, depth is treated as a discretization of continuous time. Features are thus evolved according to a vector field $F_{\theta}$ defined in terms of an algebraic bracket. This bracket arises from variational considerations and enforces structural rigidity on the propagation of information, guaranteeing useful properties such as dynamical stability and the existence of a global invariant. Depending on the desired behavior, \cite{gruber2023reversible} implements four options for the message-passing dynamics, each conferring strict guarantees on a learnable notion of energy or entropy: Hamilton's least-action principle (conservative), a generalized gradient flow (totally dissipative), a double-bracket system (partially dissipative), or a metriplectic system (thermodynamically complete). The present work employs the energy-conserving Hamiltonian bracket proposed in \cite{gruber2023reversible}, which performed best across our experiments. 

To describe precisely how this works, recall that a (noncanonical) Hamiltonian system  describes the evolution of a state variable 
$\mathbf{x}$ via Hamilton's equations:
\begin{equation}
    \dot{\mathbf{x}} = \mathbf{L}(\mathbf{x})\nabla E(\mathbf{x}),
\end{equation}
where $E$
is a Hamiltonian function (i.e., total energy) of the state and $\mathbf{L}$ is a skew-symmetric matrix field with additional structure.\footnote{Specifically, one usually enforces the Jacobi identity, a complicated PDE in the entries of $\mathbf{L}$, which is not explicitly incorporated in this parameterization.}
In the present graph setting, the Hamiltonian $E$ governs the latent node–edge feature pairs $\mathbf{x}=(\mathbf{q},\mathbf{p}) \in \mathbb{R}^{(V+E)\times N_f}$ whose components exist in feature spaces equipped with state-varying inner product matrix fields $\mathbf{A}_0(\mathbf{q})\in\mathbb{R}^{V\times V}$ and $\mathbf{A}_1(\mathbf{q})\in\mathbb{R}^{E\times E}$. These inner products play the role of Riemannian metrics on the node/edge feature spaces and are specifically chosen to incorporate the influence of graph attention, which we discuss further below.

We define the combined node-edge inner product  matrix field  $\mathbf{A}$ (for each $\mathbf{q}$) as:
\begin{equation}
    \mathbf{A}(\mathbf{x}) = \mathrm{diag}\left(\mathbf{A}_0(\mathbf{q}),\mathbf{A}_1(\mathbf{q})\right) \in \mathbb{R}^{(V+E)\times (V+E)}.
\end{equation}
It then follows that the $\mathbf{A}$-gradient\footnote{Here, we mean the gradient with respect to the derivative operator induced by the inner product involving $\mathbf{A}$.} of any function $E(\mathbf{x})\in\mathbb{R}^{V+E}$ is given by
\begin{equation}
    \nabla^A E(\mathbf{x}) = \mathbf{A}(\mathbf{x})^{-1}\nabla E(\mathbf{x}),
\end{equation} and the $\mathbf{A}$-adjoint of any linear operator $\mathbf{L}(\mathbf{x})\in\mathbb{R}^
{(V+E)\times(V+E)}$ is given by
\begin{equation}
    \mathbf{L}(\mathbf{x})^* = \mathbf{A}(\mathbf{x})^{-1}\mathbf{L}(\mathbf{x})^\intercal\mathbf{A}(\mathbf{x}),
\end{equation}
(c.f. \cite{gruber2023reversible}).  Choosing the particular expressions:
\begin{equation}
\mathbf{L}(\mathbf{x})=\begin{pmatrix}
0 & -d_0^* \\
d_0 & 0
\end{pmatrix} \quad \text{ and } \quad E(\mathbf{x})=\frac{1}{2}(\lvert\mathbf{q}\rvert^2+\lvert\mathbf{p}\rvert^2),
\end{equation}
in terms of the graph gradient or incidence matrix $d_0\in\mathbb{R}^{E\times V}$ of edges on nodes and its $\mathbf{A}$-adjoint $d_0^* = \mathbf{A}_0^{-1}d_0^\intercal\mathbf{A}_1$, it follows that the evolution equations $\dot{\mathbf{x}} = \mathbf{L}(\mathbf{x})\nabla^A E(\mathbf{x})$, or:
\[
\begin{pmatrix}
\dot{\mathbf{q}} \\
\dot{\mathbf{p}}
\end{pmatrix}
=
\begin{pmatrix}
0 & -d_0^* \\
d_0 & 0
\end{pmatrix}
\begin{pmatrix}
\mathbf{A}_0^{-1} & 0 \\
0 & \mathbf{A}_1^{-1}
\end{pmatrix}
\begin{pmatrix}
\mathbf{q} \\
\mathbf{p}
\end{pmatrix}
=
\begin{pmatrix}
- d_0^* \mathbf{A}_1^{-1} \mathbf{p} \\
d_0 \mathbf{A}_0^{-1} \mathbf{q}
\end{pmatrix},
\]
generate Hamiltonian dynamics on the latent space containing $\mathbf{x}$. This can be seen from the fact that $\mathbf{L}^* = -\mathbf{L}$ is skew-adjoint for all $\mathbf{x}$, which guarantees that the instantaneous energy rate $\dot{E}$ satisfies:
\begin{equation}
\begin{aligned}
    \dot{E}(\mathbf{x}) &= \left(\dot{\mathbf{x}},\nabla^A E(\mathbf{x})\right)_{\mathbf{A}}\\
    &= \left(\mathbf{L}(\mathbf{x})\nabla^A E(\mathbf{x}), \nabla^A E(\mathbf{x})\right)_{\mathbf{A}}\\
    &= -\left(\nabla^A E(\mathbf{x}),\mathbf{L}(\mathbf{x})\nabla^A E(\mathbf{x})\right)_{\mathbf{A}}\\
    &= 0,
\end{aligned}
\end{equation}
where $(\cdot,\cdot)_{\mathbf{A}} = \langle \cdot,\mathbf{A}\cdot\rangle$ denotes the $\mathbf{A}$-inner product on $\mathbb{R}^{V+E}$.
The algebraic bracket $\{E,F\} = (\nabla E,\mathbf{L}\nabla F)_{\mathbf{A}}$ therefore 
guarantees that information flows along level sets of the Hamiltonian $E$,
generalizing this consequence of classical mechanics to the graph setting and  ensuring that node/edge feature updates align with Hamilton's principle of least action. We emphasize that the Hamiltonian $E$ is a function on the latent features in this case, meaning that energy conservation in the original variables is not imposed. Rather, these latent Hamiltonian dynamics guarantee a global invariant on the layer-wise (i.e., discrete time) propagation, serving to stabilize information flow during message passing and prevent oversmoothing of the latent features.

It remains to explain how the learnable inner products $\mathbf{A}_0, \mathbf{A}_1$ implement graph attention. Letting $N_f>0$ denote the latent dimension of the node features (e.g, ice thickness, basal friction, etc.), 
the entries of $\mathbf{A}_1$ are defined in terms of ``query vectors'' $\mathbf{Wq}_i$ and ``key vectors'' $\mathbf{Kq}_i$ coming from learnable linear embeddings $\mathbf{W,K}\in\mathbb{R}^{N_h\times N_f}$, whose images are contained in a ``hidden feature space'' $\mathbb{R}^{N_h}$. For edge $\alpha = (i,j)$, we have:
\begin{equation}\label{eq:A1}
    \left[\mathbf{A}_1(\mathbf{q})\right]_{\alpha\alpha} = \exp\left(\mathbf{Wq}_i\cdot \mathbf{Kq}_j + \mathbf{Wq}_j\cdot \mathbf{Kq}_i\right).
\end{equation}
Notice that $\mathbf{A}_1\in\mathbb{R}^{E\times E}$ is diagonal (hence symmetric) with strictly positive entries in the space of graph edges, leading to a valid inner product on $\mathbb{R}^E$.
However, it can also be represented with nodal indices by defining a matrix field $\mathbf{a}_1(\mathbf{q})\in \mathbb{R}^{V\times V}$ with entries $\left[\mathbf{a}_1(\mathbf{q})\right]_{ij} = \left[\mathbf{A}_1(\mathbf{q})\right]_{\alpha\alpha}$, in which case it remains symmetric but contains off-diagonal terms. The nodal inner product $\mathbf{A}_0$ is then defined as the sum of incident edge weights:
\begin{equation}\label{eq:A0}
    \left[\mathbf{A}_0(\mathbf{q})\right]_{ii}=\sum_{j\in\mathcal{N}(i)}\left[\mathbf{a}_1(\mathbf{q})\right]_{ij} = \sum_{j\in\mathcal{N}(i)}\exp\left(\mathbf{Wq}_i\cdot \mathbf{Kq}_j + \mathbf{Wq}_j\cdot \mathbf{Kq}_i\right).
\end{equation}
Again, it can be seen that $\mathbf{A}_0$ is diagonal with strictly positive entries.  Moreover, it follows that the usual node-level attention (with symmetrized numerator) is recoverable via the expression $\mathrm{att}(\mathbf{q}_i,\mathbf{q}_j) = \left[\mathbf{A}_0(\mathbf{q})^{-1}\mathbf{a}_1(\mathbf{q})\right]_{ij}$, and it can be shown that the ``attention Laplacian'' $\Delta = d_0^*d_0$ satisfies the following graph attention network (GAT)-like update:
\begin{equation}
    \left[\Delta\mathbf{q}\right]_i = \sum_{j\in\mathcal{N}(i)}\mathrm{att}\left(\mathbf{q}_i,\mathbf{q}_j\right)\left(\mathbf{q}_i-\mathbf{q}_j\right).
\end{equation}
Importantly, the attention Laplacian $\Delta$ incorporates both topological information coming from the graph domain as well as metric information coming from the nodal representations, as opposed to the usual graph Laplacian which only accounts for connectivity.  
Therefore, the differential operators computed with the inner products $\mathbf{A}_0,\mathbf{A}_1$ naturally mimic key properties of graph attention while maintaining the conservation properties required for stable, long-term physical simulations.

%% file: Sections/S3_p3_dataprep.tex
\subsection{Data preparation and graph generation}\label{sec:data_prep}

Each MALI simulation generates 93 annual snapshots of ice sheet state variables. We treat each snapshot as an independent training sample, and we randomly select 20 simulations for validation and 20 for testing.

As described in \Cref{sec:ice_sheet_model}, MALI discretizes the thickness and temperature equations on a Voronoi mesh and solves the velocity equations on its dual Delaunay triangulation using low-order Lagrangian finite elements. We adopt this dual triangulation as the graph on which our GNN surrogate operates: the Delaunay nodes correspond to velocity degrees of freedom, and the FEM connectivity defines graph edges (\Cref{fig:graph}). This yields a fixed graph with 18,544 nodes and 54,962 edges.

Operating directly on this graph has two main advantages. First, it is naturally compatible with the unstructured meshes used in modern ice sheet models. Because GNNs act on node-edge structures and are agnostic to spatial ordering, no remeshing, interpolation, or other geometric preprocessing is required if the domain or mesh were to change. This stands in contrast to surrogate models based on regular grids, such as GNNs, which require substantial data transformation. Second, the graph-based formulation enables transferability: a GNN trained on one domain or mesh can be fine-tuned on a related graph. For dynamic, plastic systems such as ice sheets, where domains and meshes may evolve, this provides a major computational advantage. We return to this point in \Cref{sec:TL}.

Each node is assigned a seven-dimensional feature vector consisting of ice thickness, bed topography (or elevation), basal friction, and two Boolean indicators for grounded versus floating ice. \Cref{fig:input_features} shows a sample of these features. The model outputs are the $x-$ and $y-$ components of the velocity, which we also report in magnitude for visualization.

We normalize all scalar node and edge features using z-score normalization, except for the computed edge-wise distances, which use a min-max normalization to avoid distortion. For a given node feature $Z$, the normalized value $z'$ is
\begin{equation}
    z'=\frac{z-\mu_{\Omega}(Z)}{\sigma_{\Omega}(Z)},
\end{equation}
where $\mu_{\Omega}$ and $\sigma_{\Omega}$ denote the global mean and standard deviation of $Z$ over the full domain $\Omega$. This centers each feature and places it on a comparable scale without distorting relative differences between data points.

Geometric information enters through edge-wise distances between node coordinates. These distance features are scaled to the fixed range [0,1] using min-max normalization,
\begin{equation}
    d'=\frac{d-\min_{\Omega}d}{\max_{\Omega}d-\min_{\Omega}d},
\end{equation}
to preserve their relative magnitudes while keeping them numerically well-conditioned.

Global statistics $\mu_{\Omega},\sigma_{\Omega},\min_{\Omega}$ and $\max_{\Omega}$ are computed once and reused across all experiments, including pre-training and fine-tuning on different subgraphs, to ensure consistency. Training is performed on normalized features and targets, while all reported test metrics are computed using de-normalized velocity predictions and ground truth.

%% file: Sections/S3_p4_training.tex
\subsection{Training details}\label{sec:train}

We learn the parameter set $\theta$ 
of the GNN surrogate by minimizing the mean squared error (MSE) between predicted and simulated velocities:
\begin{equation}
    \mathcal{L}(\theta)=\frac{1}{N_{\beta}N_{\tau}}\sum_{i=1}^{N_\beta}\sum_{j=1}^{N_\tau}\sum_{\mathbf{x}\in\Omega}\big({\mathbf{u}_{\theta}}(\mathbf{x},t^j;\beta_i)-\mathbf{u}(\mathbf{x},t^j;\beta_i)\big)^2,
\end{equation}
where we train on $N_\tau$ randomly selected time steps $t^j$, $j=1, \ldots, N_{\tau}$, for $N_\beta$ independent basal friction fields $\beta_i$, $i=1, \ldots, N_{\beta}$. MSE is the standard choice for GNN regression problems and heavily penalizes large errors, which is desirable in our setting: the largest absolute errors typically occur in fast-flow regions near the glacier terminus, where accurate predictions are most critical. The encoder, decoder, internal attention and bracket parameters of the GNN are all learned as part of $\theta$. We train the network using mini-batch gradient descent on a cluster equipped with NVIDIA A100 40GB GPUs.

We deliberately do not include any physics-based regularization terms in the loss (for example, penalties enforcing mass conservation residuals). The GNN architecture itself encodes useful structural biases, although this primarily serves to stabilize information flow in the latent dynamics and only implicitly encourages physically plausible behavior. Combined with our rich feature set, this is sufficient to produce stable and accurate predictions.

\begin{figure}[hbt!]
\centering
\includegraphics[width=\linewidth]{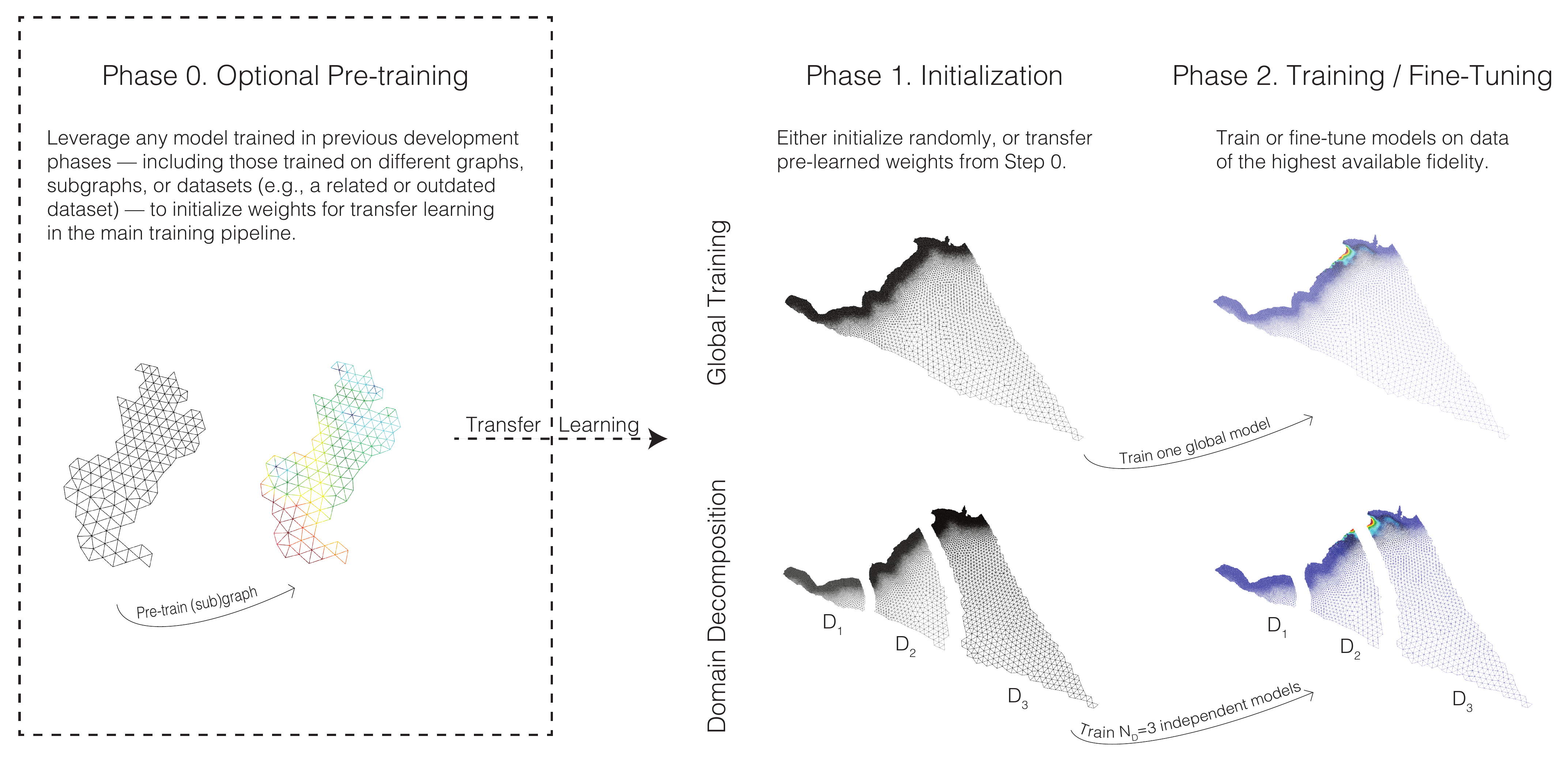}
\caption{Overview of training pipeline, with different strategies considered. Phase 0 is an optional pre-training step that utilizes work already completed during model development, or abundant data pertaining to another related graph (e.g. a subgraph of the ice sheet of interest, or a completely different ice sheet).
}
\label{fig:pipeline}
\end{figure}

The physics-inspired bracket-based GNN architecture we use, based on \cite{gruber2023reversible}, is highly configurable, allowing for different choices of brackets, ODE integrators, and encoding/decoding strategies. Our experiments indicate that architectural variations of this type are far less impactful than feature design and hyperparameter tuning (e.g. the learning rate scheduler), though they do affect training efficiency. We provide additional details on model parameters and runtimes in \Cref{sec:appendix}.
\Cref{fig:pipeline} summarizes the overall training pipeline, including the modifications that implement our transfer-learning and domain decomposiiton strategies. These techniques are discussed in detail in  \Cref{sec:TL,sec:DD}, respectively.

%% file: Sections/S3_p5_transfer_learning.tex
\section{Transfer learning}\label{sec:TL}
A key advantage of a graph-based surrogate modeling strategy is the ability to learn the attention mechanism on one part of the domain and reuse this knowledge elsewhere on the graph. This idea, commonly referred to as transfer learning or fine-tuning, can substantially reduce training time relative to training from scratch, and it enables surrogate modeling in regimes where high-fidelity data are scarce.

Transfer learning has been widely successful in improving predictive performance when only limited high-fidelity training data are available, even if the bulk of the training has been performed on coarser or lower-dimensional data \cite{propp2025transfer}. Transferring weights from previously trained models is also known to accelerate convergence on new tasks \cite{howard2024multifidelity}, even when the learning problems differ significantly. In the context of GNNs, several transfer strategies are possible, but one of the most intuitive and effective is to pre-train on low-fidelity (or auxilliary) data and then fine-tune on a smaller set of high-fidelity samples \cite{buterez2024transfer}. This is the general strategy we adopt.

At a high level, our fine-tuning procedure consists of three steps:
\begin{enumerate}
    \item \textbf{Pre-training.} Learn parameters $\theta^*_0$ on an initial graph $G_0$, which may be a subgraph of the full domain or a related graph from a different simulation setup.
    \item \textbf{Parameter transfer.} Initialize a new model on graph $G_1$. All learnable parameters $\theta$ (encoder, decoder, and attention/bracket weights) are initialized from the pre-trained model on $G_0$. If $G_1$ differs from $G_0$, (e.g. a different mesh), all graph-dependent operators (e.g. incidence matrices $d_0$, Laplacians, etc.) are recomputed from the topology of $G_1$.
    \item \textbf{Fine tuning.} Continue training the new model with parameters $\theta_1$ on $G_1$ using the available high-fidelity data.
\end{enumerate}

This workflow can be applied whether the pre-trained model lives on a different graph or on the same graph but with different data (e.g. with a different mesh resolution or a different representation of the basal friction field). Transfer learning is therefore particularly attractive in settings where the training data may be improved over time, for example, as new forward models, inversion algorithms, or observations become available. A model initially trained on outdated or coarse data can still be extremely valuable: it provides a strong initialization that both accelerates training on the new data and reduces the amount of additional data required for training.

Operationally, step 2 above involves extracting the learnable attention functions $\mathbf{A}_0$ and $\mathbf{A}_1$ (see \eqref{eq:A1} and \eqref{eq:A0} in \Cref{sec:BracketGraphs}) from the pre-trained model and using them, together with the encoder and decoder weights, to initialize the new model. Step 3 then fine-tunes this model on the target graph. In principle, one could restrict the update to a low-rank correction or to a subset of layers; in this work, we allow all parameters to adapt during the fine-tuning stage. \Cref{fig:pipeline} illustrates one concrete realization of this strategy, in which we pre-train on a subdomain of the glacier and then use the resulting weights to provide a ``warm start'' for training a global model or additional subdomain models.

%% file: Sections/S3_p6_domain_decomposition.tex
\section{Domain decomposition}\label{sec:DD}
A key contribution of this work is the incorporation of domain decomposition (DD) into the GNN training pipeline as a mechanism for improving both computational efficiency and accuracy. At a high level, our DD approach consists of partitioning the full domain into subdomains, training an independent model on each, and then aggregating the independent subdomain predictions. DD has been extensively studied in classical numerical analysis \cite{dolean_introduction_2015,quarteroni_domain_1999,smith_domain_2004,toselli_domain_2005}. However, DD for neural networks, and for GNNs in particular, is still in its relative infancy. Recent work has started to blend DD and machine learning \cite{heinlein_combining_2021,klawonn_machine_2024}, 
including both DD for ML and ML-enhanced DD, but the consideration of GNNs is restricted to the learning of parameters for classical DD solvers. Similarly, \cite{taghibakhshi_mg-gnn_2023} use GNNs to learn DD parameters, but does not consider DD as a tool for training GNN surrogates themselves. 
The discrete nature of graph-based learning, combined with the flexibility of attention mechanisms, suggests that DD for GNNs offers unique potential.

In classical numerical simulations, DD yields iterative methods that can be viewed as a hybrid between basic iterative solvers and direct solvers. For large problems, a direct solve becomes infeasible due to superlinear growth in computational and memory cost. Standard iterative methods avoid this cost but often suffer from deteriorating convergence as the global problem size increases. DD methods strike a balance: each iteration splits the problem into subdomain solves,where direct or approximate solvers can be used efficiently, then couples them via coupling at the interfaces or overlap between the subdomains. This yields better convergence than basic iterative schemes and, with an appropriate coarse level, can even achieve numerical scalability, where iteration counts are essentially independent of global problem size. Crucially, DD localizes the computational work, enabling parallel scalability on modern high-performance architectures.

Localization is also one of the main benefits of DD in the context of neural networks. Localization can arise through the model architecture --- for example, via mixtures of experts or locally connected networks \cite{dolean_multilevel_2023,heinlein_domain_2025,howard_finite_2024,jagtap_conservative_2020,karniadakis_extended_2020,moseley_finite_2021,shang_overlapping_2025} --- or through the data, as in convolutional models trained on image patches \cite{gu_decomposition_2022,klawonn_domain_2023,pelzer_scalable_nodate,verburg_ddu-net_2025}. Beyond computational gains, localization has an interesting effect on learning dynamics. Neural networks exhibit a spectral bias, or frequency principle \cite{rahaman_spectral_2019}: high-frequency functions are typically learned more slowly than low-frequency ones, in part due to the global nature of commonly used activation functions \cite{hong_activation_2022}. By decomposing the domain, DD effectively localizes these functions, allowing high-frequency behavior to be represented and learned within smaller subproblems. This can help mitigate spectral bias \cite{dolean_multilevel_2023,moseley_finite_2021}.

In the present work, we employ a combination of model and data decomposition. On the data side, we decompose the domain in the form of the underlying graph (mesh), partitioning the ice sheet into subgraphs. On the model side, we train localized GNN surrogates on each subgraph and then aggregate their predictions into a global field. In this respect, our work shares some characteristics with localized CNN approaches for image decomposition \cite{klawonn_domain_2023}, where a specialized model is trained on each part of a decomposed image. However, while \cite{klawonn_domain_2023} considers an image classification task using CNNs and aggregates submodel predictions into a single label using another model, we address a node-wise regression problem on an irregular graph. Other related works \cite{pelzer_scalable_nodate,verburg_ddu-net_2025} for CNNs learn a single global model via weight sharing, whereas our approach explicitly trains separate local surrogates and combines them at inference time. 

These design choices motivate the developments in the following subsections. Section \Cref{sec:partitions} describes how we construct physically coherent subdomains through a spectral clustering-based partitioning of the graph, and \Cref{sec:trainDD} details how we train and aggregate the subdomain models to obtain a consistent global prediction.

\subsection{Partitioning of the graph}\label{sec:partitions}
A central challenge in DD is how to partition the domain. In classical methods, this is typically done using geometric information or structure, or via graph partitioning algorithms~\cite{karypis_fast_1998,devine2002zoltan}. In the context of NNs, this topic still remains much less explored. Although recent work, e.g. \cite{pmlr-v190-trask22a}, offers some guidance, the question of how best to partition a graph for DD on GNNs remains relatively unexplored. For our purposes, an effective partition must satisfy three criteria: subdomains must be contiguous, approximately balanced in size, and aligned with meaningful physical variation in the data. Although we leave a thorough treatment of this topic for future work, we make a first step towards addressing it using a spectral-clustering approach that blends spatial, feature, and target similarity with a modified $k$-means procedure to encourage approximately balanced, physically coherent subdomains.

\begin{figure}[hbt!]
\centering
\includegraphics[width=0.4\linewidth]{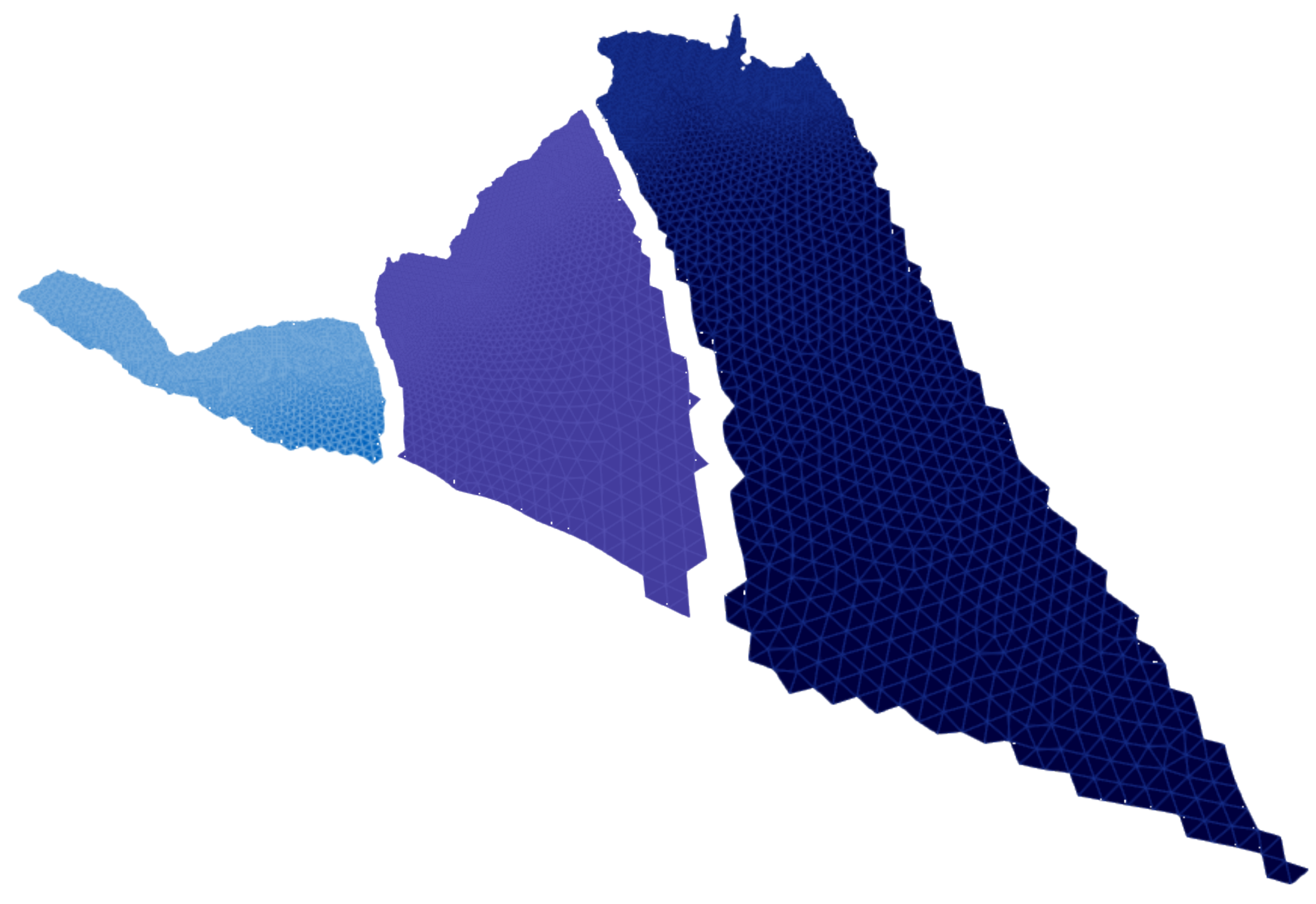}
\caption{Subdomains produced by spectral clustering algorithm, for $k=3$. }
\label{fig:spectral_partition}
\end{figure}

We partition the mesh by first building a symmetric weighted adjacency matrix $W$ on the existing mesh edges so that contiguity is preserved and only neighboring cells interact. We borrow from the field of computer vision, and the work of \cite{shi2000normalized} in particular, weighting each edge $(u,v)$ with a Gaussian similarity that combines spatial proximity with similarity in both node features and target values, yielding higher affinity for nearby nodes with similar covariates and velocities. Concretely, letting $\Delta x_{uv}\in\mathbb{R}^2$ be the difference in spatial coordinates, $\Delta f_{uv}\in\mathbb{R}^3$ be the difference in features (thickness, friction, and bed topography), and $\Delta y_{uv}\in\mathbb{R}^2$ be the difference in velocity components (all computed on averages across training realizations), the resulting edge weight is:
\begin{equation}
    W_{uv}=\exp\Big(\frac{\norm{\Delta x_{uv}}}{\sigma_x^2}\Big)\exp\Big(\frac{\norm{\Delta f_{uv}}}{\sigma_f^2}\Big)\exp\Big(\frac{\norm{\Delta y_{uv}}}{\sigma_y^2}\Big),
\end{equation}
yielding a sparse $W$ supported on mesh edges.
From $W$, we form the weighted graph Laplacian $L=D-W$ and compute its smallest nontrivial eigenvectors, which provide a spectral embedding in which Euclidean distances reflect relaxed cut objectives \cite{ng2001spectral}. We then cluster the embedded points with a size-penalized $k$-means algorithm that softly balances subdomain sizes by adding a small bias to each cluster's squared distance proportional to its deviation from $N/k$ points. This procedure promotes spatial coherence (because support is restricted to mesh edges), aligns boundaries with covariate and velocity structure, and avoids highly imbalanced partitions. In this work, we proceed with the clustering produced with $k=3$, shown in \Cref{fig:spectral_partition}. Importantly, the subdomain boundaries of this partition roughly align with the direction of the velocity gradient, and ensure that each partition exhibits suitable variation in each feature.

Although overlapping subdomains are often beneficial in the classical DD setting, where they allow information about the PDE's nullspace to propagate across interfaces, we did not observe similar advantages in our setting. Prior work on DD with CNNs required either explicit overlap or a coarse global model to mitigate interface effects and ensure consistency across subdomains. In contrast, our GNN surrogate remains remarkably robust even with non-overlapping partitions: we do not see error growth near subdomain boundaries, nor do we observe the kinds of interface discontinuities one might expect when models are trained independently. This suggests that, because our GNN surrogate is fully data-driven and optimized under an MSE objective rather than solving a PDE directly, there is no analogous need to transport nullspace information across subdomains. We therefore adopt non-overlapping partitions based on mesh elements, which simplifies the training pipeline while retaining accuracy. Overlapping partitions could still be useful in regimes with extremely small neighborhoods or when the receptive field is severely truncated at partition boundaries, but such issues did not arise in our experiments. 

\begin{algorithm}[t]
\caption{Domain-decomposed training of $N_D$ sub-domain GNNs}\label{alg:dd-gnn}
\SetKwInOut{Input}{Require}
\SetKwInOut{Output}{Ensure}

\Input{
  Partitioned training sets $\{\mathcal D_i^{\text{train}}\}_{i=1}^{N_D}$; \\
  Global validation and test sets $\mathcal D^{\text{val}}$ and $\mathcal D^{\text{test}}$; \\
  Normalization statistics $(\mu_x, \sigma_x, \mu_y, \sigma_y)$; \\
  Hyper-parameters $\mathrm{opt}$ (epochs, learning rate, warm start, \dots)
}
\Output{
  Trained weights $\{\Theta_i\}_{i=1}^{N_D}$, loss history, best global
  prediction on $\mathcal D^{\text{val}}$
}

\For{$i \gets 1$ \KwTo $N_D$}{
  $\mathcal D_i^{\text{train}} \gets 
      \textsc{NormalizeZ}(\mathcal D_i^{\text{train}};
      \mu_x,\sigma_x,\mu_y,\sigma_y)$\;
  Initialize $\Theta_i$ (optionally from pre-trained weights)\;
}
Normalize validation inputs (z-score):  
$\mathbf X^{\text{val}} \gets 
   \textsc{NormalizeZ}(\mathbf X^{\text{val}}; \mu_x,\sigma_x)$\;
Define node-wise weights $w_{i,v}$ (for overlaps)\;

\For{$e \gets 1$ \KwTo $\mathrm{opt.num\_epochs}$}{
  \For{$i \gets 1$ \KwTo $N_D$}{
    \textsc{TrainEpoch}$(\Theta_i, \mathcal D_i^{\text{train}})$\;
  }

  Set aggregate prediction $\hat{\mathbf Y} \gets 0$ and
  accumulation weights $\mathbf W \gets 0$\;

  \For{$i \gets 1$ \KwTo $N_D$}{
    $\hat{\mathbf Y}_i \gets \textsc{Predict}
        (\Theta_i, \mathbf X^{\text{val}}[\text{subgraph}_i])$\;
    Undo normalization:
    $\hat{\mathbf Y}_i \gets 
        \hat{\mathbf Y}_i^{(z)} \,\sigma_y + \mu_y$\;
    Add to aggregate:
    $\hat{\mathbf Y}[\text{subgraph}_i] \gets
       \hat{\mathbf Y}[\text{subgraph}_i] + w_{i,v}\,\hat{\mathbf Y}_i$\;
    Update weights:
    $\mathbf W[\text{subgraph}_i] \gets
       \mathbf W[\text{subgraph}_i] + w_{i,v}$\;
  }

  Weighted averaging for overlaps:
  $\hat{\mathbf Y} \gets \hat{\mathbf Y} \oslash \mathbf W$ (element-wise)\;
  Evaluate global validation loss; if improved, save $\{\Theta_i\}$ and log metrics\;
}

After training, report final performance on held-out test set $\mathcal D^{\text{test}}$\;

\end{algorithm}

\subsection{Training with subdomains}\label{sec:trainDD}
Having identified $N_D$ suitable subdomains, we train $N_D$ independent GNN surrogates using the procedure described in \Cref{alg:dd-gnn}. Training of the subdomain models closely mirrors the global training strategy described in \Cref{sec:train}, with three main differences: (i) reduced memory cost per subgraph, (ii) opportunities for parallelization across subdomains, and (iii) the need to aggregate subdomain predictions together at inference time.

For non-overlapping partitions, where each node belongs to exactly one subdomain, the aggregation step is trivial: the global prediction is obtained by placing each subdomain's predictions onto its corresponding portion of the full graph, with no further processing required. For overlapping partitions, however, multiple subdomain models may produce predictions for the same node. In this case, we assign each node $v$ a set of weights $\{w_{i,v}\}$, where $w_{i,v}$ specifies the contribution of subdomain $i$ to the final aggregated prediction at node $v$. If $\hat{y}_{i,v}$ denotes the prediction from subdomain $i$ at node $v$, the aggregated prediction is:
\begin{equation}
    \hat{y}_v=\frac{\sum_{i=1}^{N_D}w_{i,v}\hat{y}_{i,v}}{\sum_{i=1}^{N_D}w_{i,v}}.
\end{equation}
Setting $w_{i,v}=1$ for all contributing subdomains yields a simple average over the predictions associated with node $v$. In the non-overlapping case, only a single subdomain contains node $v$, so the formula reduces to $\hat{y}_v=\hat{y}_{i,v}$; thus the aggregation step is entirely non-intrusive.
Other weighting choices are possible, however, and could be designed to emphasize particular subdomains near boundaries or reflect local uncertainty estimates, for example. \Cref{fig:pipeline} presents a schematic overview of the domain-decomposed training and aggregation process for the case of $N_D=3$ non-overlapping subdomains. 

%% file: Sections/S4_results.tex
\section{Results}\label{sec:results}
\begin{figure}[hbt!]
\centering
\includegraphics[width=0.9\linewidth]{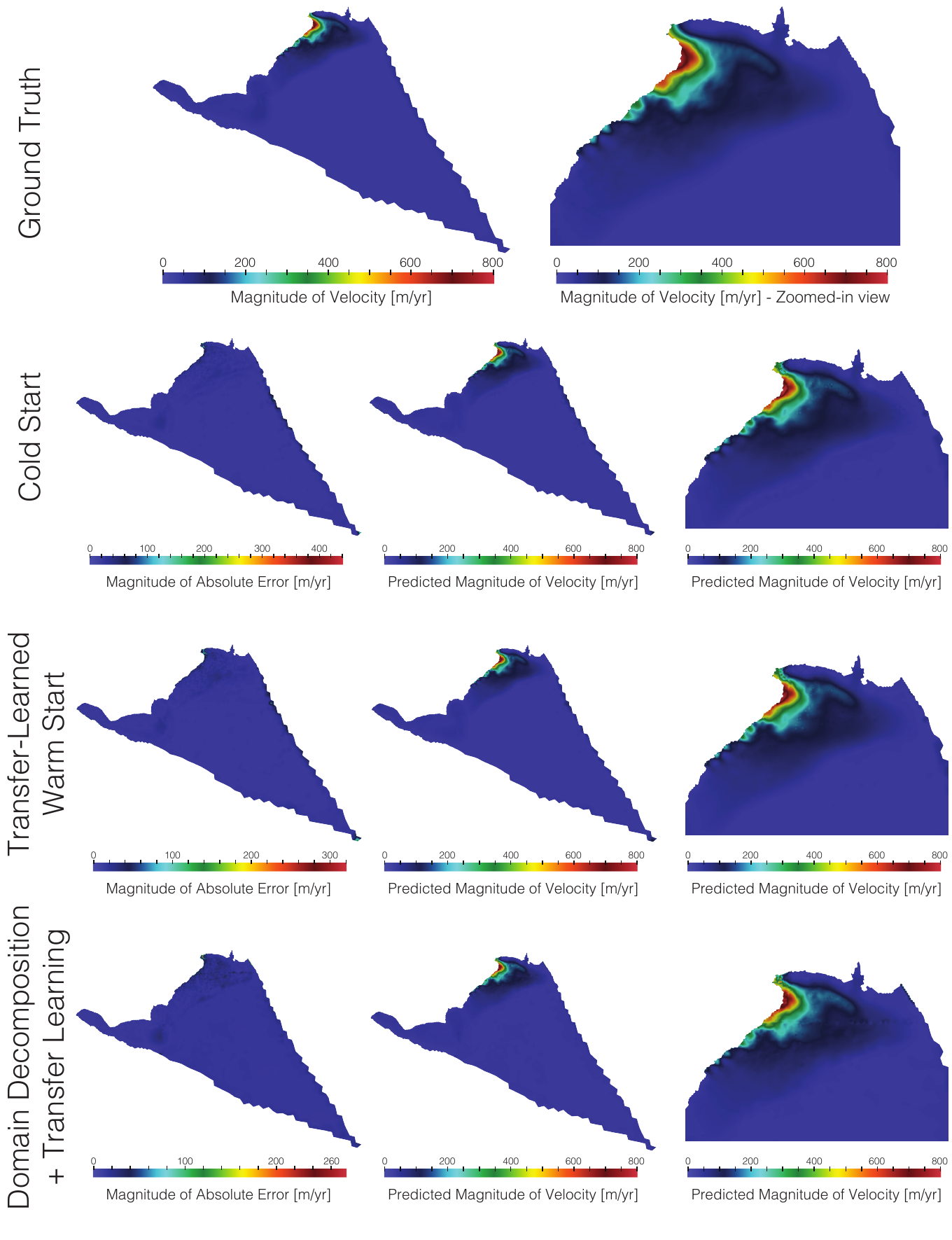}
\caption{Comparison of three training strategies on the same test snapshot. Top row: ground-truth velocity magnitude (full domain and zoomed terminus view). Rows 2-4 show: (i) cold start, (ii) warm start (pre-trained on a subgraph and fine-tuned globally), and (iii) warm start plus DD, where subdomain models are pre-trained, fine-tuned, and stitched at inference. Columns display the magnitude of pointwise velocity error, full-field velocity prediction, and zoomed-in velocity prediction. The error color scale is intentionally not fixed across rows to highlight improvements: the overall magnitude of error decreases substantially from cold start to warm start to warm start plus DD, demonstrating progressively stronger accuracy, particularly in the terminus region.
}
\label{fig:results_ALL}
\end{figure}

We evaluate the performance of our physics-inspired GNN surrogate and investigate how transfer learning and domain decomposition influence training efficiency and predictive accuracy. A high-level visual overview of our training strategies (cold start, warm start, and warm start combined with DD) is provided in \Cref{fig:pipeline}. \Cref{fig:results_ALL} presents direct, side-by-side qualitative comparisons of their predictions on the same test snapshot.

We begin with the most challenging case as a baseline: training a single global model from scratch (``cold start''). With 25 basal friction fields and 40 snapshots per field (1000 samples total), the model successfully captures the large-scale structure of the velocity field. Even in this data-limited regime, errors concentrate near the grounding line and fast-flow terminus region, where velocities are highest and gradients are steepest. This can be seen by comparing the first and second rows in \Cref{fig:results_ALL}.

\begin{figure}[hbt!]
\centering
\includegraphics[width=0.8\linewidth]{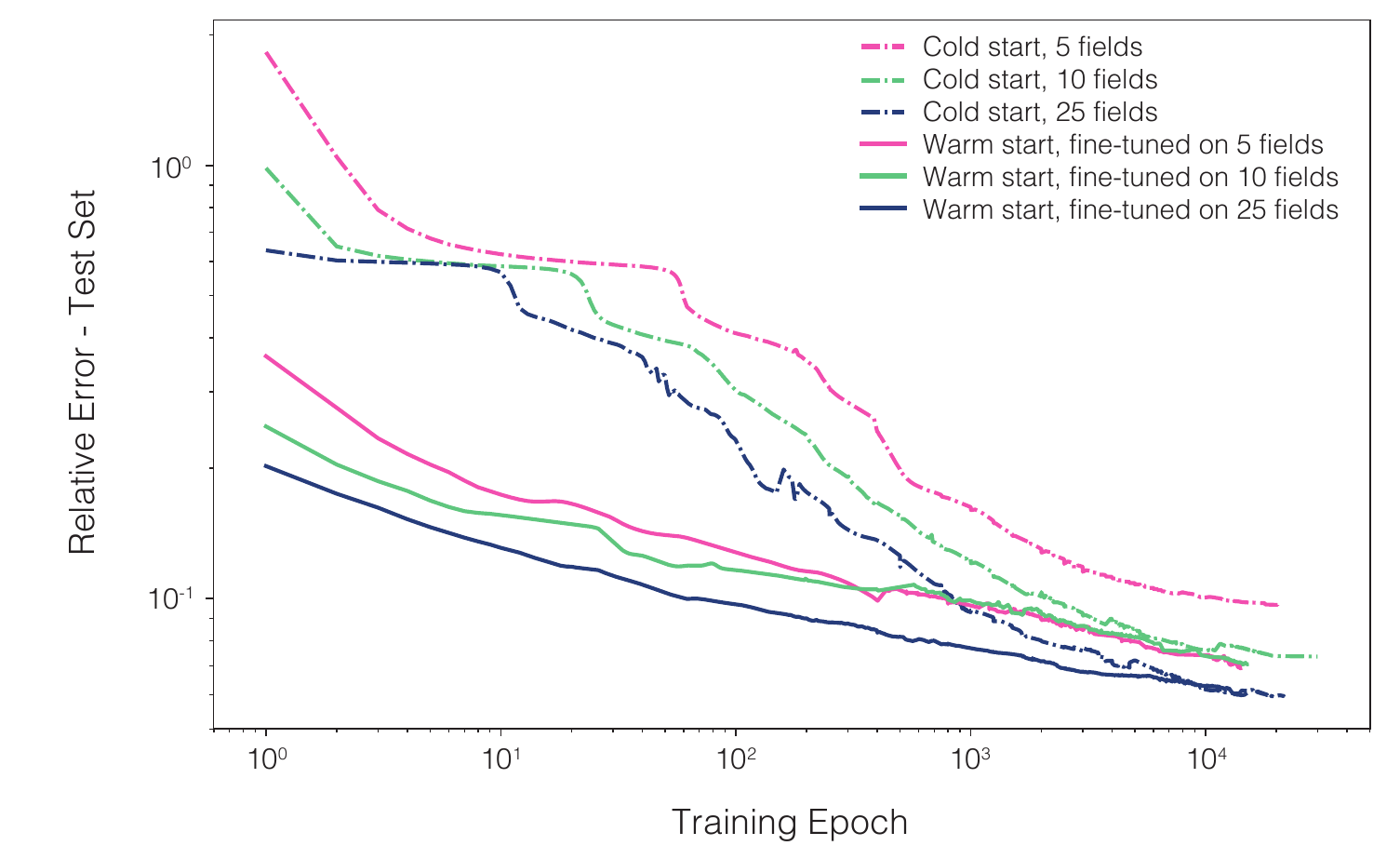}
\caption{Relative test error for cold start and warm start strategies, trained on different numbers of basal friction fields. Warm start indicates fine-tuning from a model pre-trained on 40 friction fields for the southern interior region of the domain. Fine-tuning consistently accelerates training and lowers error compared to training from scratch.
}
\label{fig:results_fine_tuned}
\end{figure}

Next, we assess ``warm starts,'' where a model is pre-trained on a subgraph and then fine-tuned globally. As shown in \Cref{fig:results_fine_tuned}, pre-training consistently accelerates convergence and reduces or maintains error across all data-regime conditions. When data are scarce (e.g. only 5-10 basal friction realizations), fine tuning provides particularly large gains. Compared to the cold start strategy, transfer learning plus fine-tuning achieves lower test error under both constrained training time (fixed epochs\footnote{This can be seen by comparing the error achieved by the dashed and solid lines at a given epoch.}) and constrained data (fixed training samples\footnote{This can be seen by comparing dashed and solid lines of the same color.}). The second row of \Cref{fig:results_ALL} demonstrates that error magnitude decreases markedly relative to the cold-start model, especially near the terminus.

\begin{figure}[hbt!]
\centering
\includegraphics[width=0.8\linewidth]{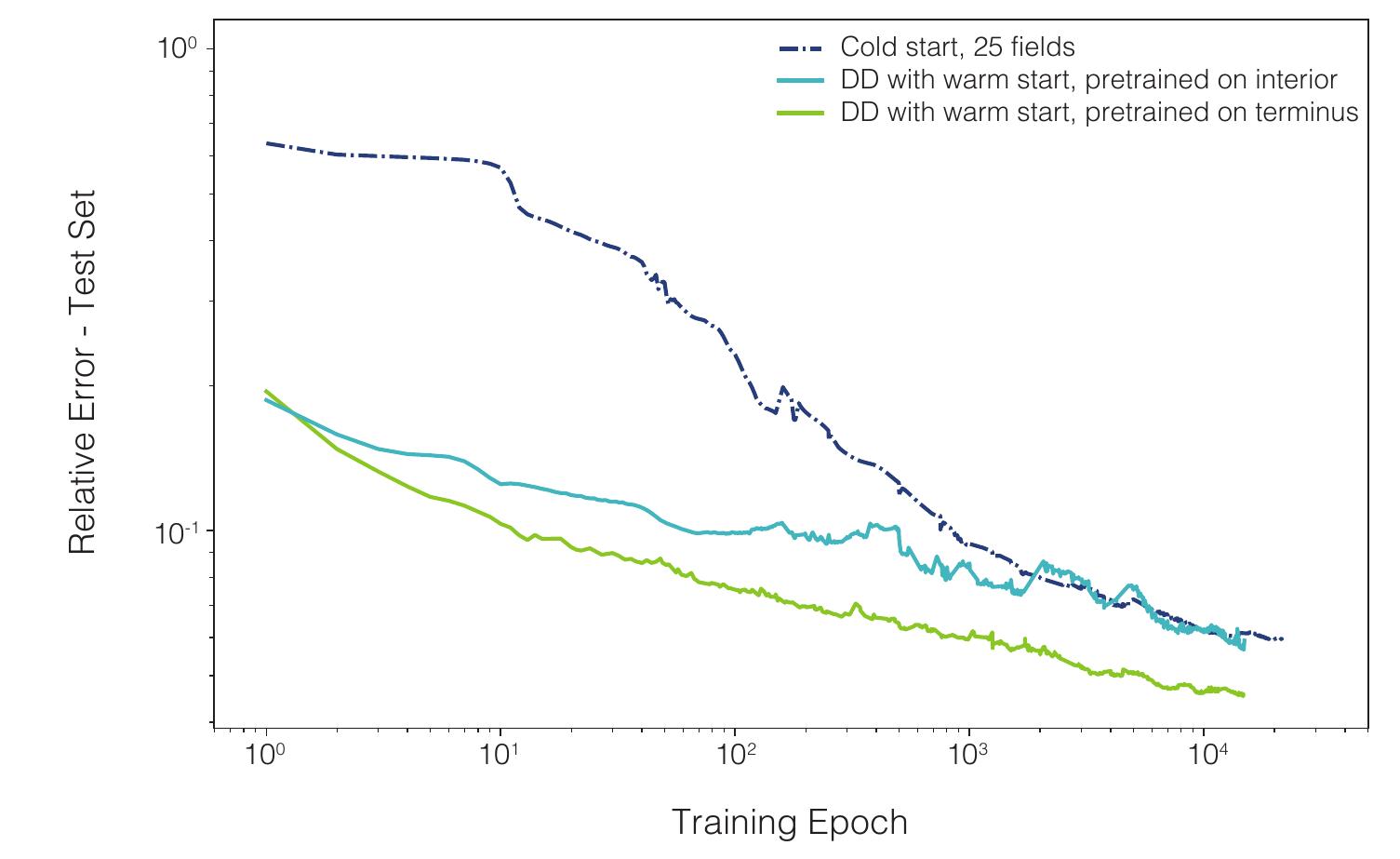}
\caption{Relative test error for domain decomposition (DD) with warm start compared to global cold start. Pre-training on the terminus region (northeastern subgraph) yields faster convergence and lower error than pre-training on the interior (southernmost subgraph).
}
\label{fig:results_DDTL}
\end{figure}

Finally, we combine transfer learning with domain decomposition. We partition the domain into subgraphs, pre-train on one region, and then transfer the learned parameters into the remaining subdomain models before fine-tuning. This strategy yields the fastest convergence (\Cref{fig:results_DDTL}) and the highest predictive accuracy (\Cref{fig:results_ALL}, bottom row) of all methods we tested. Predictions from this combined transfer learning and DD strategy (\Cref{fig:results_ALL}) show that errors near the terminus are nearly eliminated, with strong predictive accuracy even in regions of high velocity. The biggest improvements arise when pre-training uses the northeastern terminus region (rightmost subdomain in \Cref{fig:spectral_partition}), where velocity variability is greatest; initializing subdomain models from this high-complexity region transfers richer local structure and produces notably improved performance, as seen in \Cref{fig:results_DDTL}.

In summary, while all strategies eventually produce qualitatively correct predictions, transfer learning --- and especially transfer learning combined with DD --- dramatically improves both efficiency and accuracy. These results demonstrate the substantial computational savings and performance gains obtained by leveraging graph structure, localized dynamics, and warm-start initialization in large-scale surrogate modeling.

\clearpage

%% file: Sections/S5_conclusion.tex
\section{Discussion}\label{sec:discussion}

We have presented a framework for scalable surrogate modeling of large-scale PDE-governed systems, motivated by the pressing need for efficient yet reliable projections of complex physical phenomena. Our approach combines GNNs, transfer learning, and domain decomposition to address the challenges inherent in training surrogates on large, unstructured meshes. By partitioning complex domains into manageable subgraphs, we enable training at scale while maintaining stability and physical accuracy.

Our results demonstrate that domain decomposition offers both computational and modeling benefits. Training on subdomains reduces memory requirements and training time, but it also provides a structural advantage: each subdomain forms a natural unit for knowledge transfer. Information learned in one region of the domain can be re-used and adapted in others, accelerating convergence and improving generalization. 
This perspective aligns with multifidelity learning, where inexpensive or partially aligned data sources are leveraged to boost performance on expensive tasks. In settings such as ice sheet modeling, where high-fidelity simulations remain prohibitively expensive for uncertainty quantification, these strategies offer a path toward practical, data-efficient surrogates.

Together, these elements lay the groundwork for a new generation of graph-based surrogates for large-scale physical systems. By exploiting domain structure and physics-inspired latent dynamics, we move closer to ML models that are not only computationally efficient but also scientifically trustworthy. More broadly, our results suggest that graph-based domain decomposition is a practical and versatile tool for building surrogates that remain accurate across heterogeneous regions, changing meshes, and evolving modeling objectives.

\subsection{Towards foundation models for uncertainty quantification}
\begin{figure}[h!]
\centering
\includegraphics[width=0.9\linewidth]{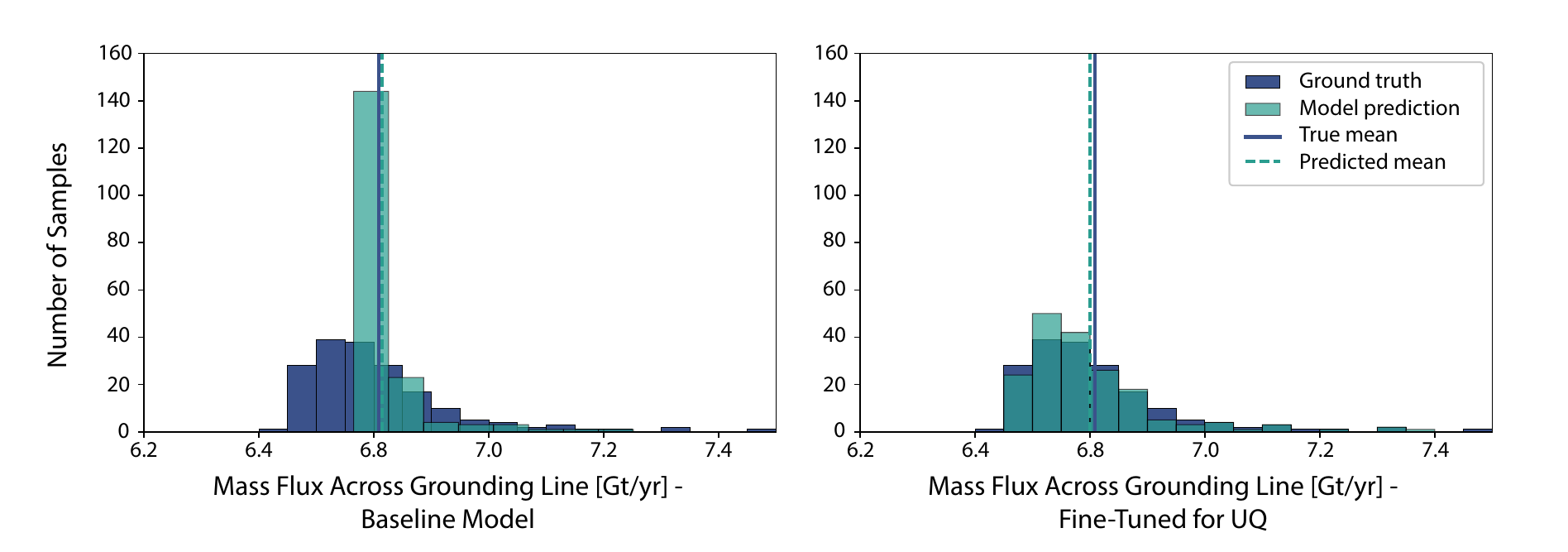}
\caption{Histograms of the distribution of ice mass flux across grounding line at $t=60$ years across 180 realizations of basal friction. Left: results using baseline high-performing model. Right: results after fine-tuning on curated UQ training set.}
\label{fig:UQ_teaser}
\end{figure}

A particularly promising direction, highlighted by our preliminary experiments on grounding-line flux, is the use of general-purpose surrogate models that can be lightly fine-tuned for downstream UQ tasks. When trained on many distinct basal-friction fields, our model learns a representation that captures the average velocity response across basal friction conditions, yielding accurate full-field predictions. This average-case accuracy emerges naturally from training on a shared latent structure across realizations. However, when the goal shifts from predicting mean behavior to capturing the variability induced by uncertain inputs, the baseline model alone may be insufficient. As illustrated by our grounding-line flux example (\Cref{fig:UQ_teaser}, left), the untuned surrogate model correctly reproduces the true mean grounding line flux but fails to capture the spread across basal friction realizations: the predicted distribution is too narrow, even though the mean is correct. 

Introducing a targeted fine-tuning phase changes this picture. When we continue training on a small dataset in which each sample is a single snapshot drawn from a different basal-friction realization (rather than our usual training sets that include many time steps from the same realization), the surrogate not only maintains an accurate estimate of the mean flux but also closely recovers the full distribution induced by basal-friction uncertainty (\Cref{fig:UQ_teaser}, right). These results suggest that fine-tuning on data designed to emphasize variability in the input space may encourage the model to pick up UQ-relevant sensitivity directions without retraining from scratch.

These preliminary results indicate that foundation-model behavior is attainable in this context: a pre-trained, domain-decomposed surrogate can be efficiently adapted to a new UQ objective with only modest additional data and computation. Looking ahead, such capabilities could bring ice sheet modeling closer to the ``foundation model'' paradigms emerging in other scientific domains, where a single, broadly trained surrogate can be rapidly specialized for new scientific questions, forcing scenarios, or UQ objectives. Developing principled strategies for the design of fine-tuning datasets and objectives for UQ is an important focus of ongoing and future work.

\section*{Acknowledgments}

This work performed in part at Sandia National Laboratories and Pacific Northwest National Laboratory was supported by the U.S. Department of Energy, Office of Science, Office of Advanced Scientific Computing Research through the SEA-CROGS project (PNNL Project No. 80278).
Sandia National Laboratories is a multimission laboratory managed and operated by National Technology \& Engineering Solutions of Sandia, LLC, a wholly owned subsidiary of Honeywell International Inc., for the U.S. Department of Energy’s National Nuclear Security Administration under contract DE-NA0003525. Pacific Northwest National Laboratory is a multi-program national laboratory
operated for the U.S. Department of Energy by Battelle Memorial Institute under Contract No. DE-AC05-76RL01830. This paper describes objective technical results and analysis. Any subjective views or opinions that might be expressed in the paper do not necessarily represent the views of the U.S. Department of Energy or the United States Government.

%% file: Sections/S6_appendix.tex
\section{Additional model details}\label{sec:appendix}

In Table~\ref{tab:gnn_params}, we summarize the architecture and hyperparameter choices for the BracketGraph GNN used as our surrogate model. Overall, we found the model to be relatively insensitive to many of these settings. For instance, increasing the number of attention heads from 2 to 4 or the hidden dimension from 20 to 50 produced results that were indistinguishable from the defaults reported here. By contrast, shrinking the encoder and decoder widths from 32 to 16 approximately doubled the final relative error, indicating greater sensitivity to these layers.

With the configuration in Table~\ref{tab:gnn_params}, the “off-line’’ training cost of the surrogate is roughly 30 hours. This can be reduced by employing explicit integration schemes such as forward Euler, although at the expense of some loss in accuracy. Once the surrogate is trained, inference takes approximately 15 ms per sample.

\begin{table}[hbt!]
\centering
\caption{Graph neural network (GNN) architecture and hyperparameters.}
\label{tab:gnn_params}
\begin{tabular}{l c}
\toprule
\textbf{Parameter} & \textbf{Value} \\
\midrule
Integrator method &  Implicit Adams-Bashforth-Moulton \\
Bracket type  &  Hamiltonian \\
Input node feature dimension ($d_\text{in}$) & 5 \\
Hidden dimension ($d_\text{hid}$) & 20 \\
Output feature dimension ($d_\text{out}$) & 2 \\
Number of Neural ODE timesteps &  2\\
Message-passing encoder/decoder width & 32 \\
Attention heads & 2 \\
Optimizer &  Adam\\
Learning rate & 1e-3 \\
Gamma & 0.95 \\
Step size & 250 \\
\bottomrule
\end{tabular}
\end{table}

\clearpage